\documentclass[conference]{IEEEtran}
\IEEEoverridecommandlockouts
\pdfoutput=1
\usepackage{cite}
\usepackage{amsmath,amssymb,amsfonts}
\usepackage{algorithmic}
\usepackage{graphicx}
\usepackage[table,xcdraw]{xcolor}
\usepackage{textcomp}
\usepackage{xcolor}
\usepackage{marvosym}
\usepackage{hyperref}

\usepackage{multirow}
\usepackage{url}

\def\BibTeX{{\rm B\kern-.05em{\sc i\kern-.025em b}\kern-.08em
    T\kern-.1667em\lower.7ex\hbox{E}\kern-.125emX}}
\begin{document}

\title{Deep Video Anomaly Detection: Opportunities and Challenges\\
}
\author{\IEEEauthorblockN{Jing Ren\IEEEauthorrefmark{1},
        Feng Xia\IEEEauthorrefmark{1}, Yemeng Liu\IEEEauthorrefmark{2} and
        Ivan Lee\IEEEauthorrefmark{3}\textsuperscript{(\Letter)}}
    \IEEEauthorblockA{ \IEEEauthorrefmark{1}School of Engineering, IT and Physical Sciences, Federation University Australia\\
        \IEEEauthorrefmark{2}School of Software, Dalian University of Technology \\
        \IEEEauthorrefmark{3}STEM, University of South Australia \\
\href{mailto:ivan.lee@unisa.edu.au} {ivan.lee@unisa.edu.au}\\
  }
}
    

\maketitle

\begin{abstract}
Anomaly detection is a popular and vital task in various research contexts, which has been studied for several decades. To ensure the safety of people's lives and assets, video surveillance has been widely deployed in various public spaces, such as crossroads, elevators, hospitals, banks, and even in private homes. Deep learning has shown its capacity in a number of domains, ranging from acoustics, images, to natural language processing. However, it is non-trivial to devise intelligent video anomaly detection systems cause anomalies significantly differ from each other in different application scenarios. There are numerous advantages if such intelligent systems could be realised in our daily lives, such as saving human resources in a large degree, reducing financial burden on the government, and identifying the anomalous behaviours timely and accurately. Recently, many studies on extending deep learning models for solving anomaly detection problems have emerged, resulting in beneficial advances in deep video anomaly detection techniques. In this paper, we present a comprehensive review of deep learning-based methods to detect the video anomalies from a new perspective. Specifically, we summarise the opportunities and challenges of deep learning models on video anomaly detection tasks, respectively. We put forth several potential future research directions of intelligent video anomaly detection system in various application domains. Moreover, we summarise the characteristics and technical problems in current deep learning methods for video anomaly detection.
\end{abstract}


\section{Introduction}
With the decreased cost of deploying surveillance cameras, the application of video surveillance is widely expanded into different scenarios. These surveillance cameras constantly produce abundant surveillance videos, which brings a huge workload and a series of technical challenges for anomaly detection in different contexts~\cite{ramachandra2020survey}. Over the past decades, deep learning has achieved great success and showed superior performance in many tasks that were previously considered to be computationally unattainable, such as face matching~\cite{Ren2021TETCI}, recommendation system~\cite{Wan2020TNSE}, and anomaly detection~\cite{Wan2020TIIdetection}. Correspondingly, more and more efforts have been devoted to video anomaly detection with deep learning-based models, which also made great achievements in recent years~\cite{pang2021deep}.

An intelligent video anomaly detection system is capable of detecting the abnormal behaviours or entities that diverge significantly from the normality, such as identifying multiple moving objects with limited prior knowledge for video surveillance \cite{wong2017track}, or detecting specific incidents such as fighting, stampede, traffic accident, and vagrant~\cite{Kong2019ITJ,Kong2018IACCESShuman}. Video anomaly is usually contextual and defined as per the real scenarios. For example, it is normal to observe crowd gathering in supermarkets or vocal concert while abnormal when social distance is required to stall the spread of virus. Among most video anomaly detection algorithms, most of them can localise the anomalies temporally and spatially~\cite{xu2019video}. Specifically, detection process concentrates on identifying the video fragments that contain anomalies among all videos, while localization devotes to determining which frame is anomalous and explaining which part of this frame is considered anomalous. Recent relevant research can deal with both problems with deep learning-based models offering an end-to-end solution.

Due to its great importance, recent years have witnessed an upsurge of research interests and applications of intelligent video anomaly detection system. However, anomaly detection in video surveillance still faces a series of challenges:
\begin{itemize}
    \item \textbf{Ambiguousness.} Anomaly detection is broadly regarded as the process of detecting events that are not expected to appear in a specific context. However, in real-world situations, the boundary between normal and abnormal items are not partitioned clearly. For example, some normal samples will also exhibit strange characteristics that abnormal events hold, which impedes the detection accuracy of models.
    \item \textbf{Dependency.} Up to now, there has not been a unified definition of anomaly despite that it is introduced in numerous literatures. On the other hand, all these definitions could not be directly applied into a specific anomaly detection task. Even the same event are likely to have different characteristics and vary widely in different backgrounds. The contextual dependency of anomaly make the detection models not adaptable.
    \item \textbf{Sparcity and Diversity.} Different from the general classification tasks, positive samples (i.e. anomaly) are much less than negative samples in real-world anomaly detection datasets. This kind of data imbalance characteristic make the supervised models difficult to train. Besides, real-world anomalous behaviours are diverse and cannot be illustrated entirely, sometimes it may even have not happened yet. Therefore, it is impractical to consider all types of possible anomalies in one model.
    \item \textbf{Privacy.} When detecting anomalies in non-video datasets, the private information of users (e.g. name) could be replaced by random generalised codes, which has no influence on the final experimental results. While in video surveillance data especially including facial and behavioural information, the individual privacy will be offended if the data is open source. This privacy characteristic leads to the fact of being short of open-source dataset.
    \item \textbf{Noise.} With the wide coverage of video surveillance, cameras are deployed for improving safety, and they are frequently found in places such as elevators, cross roads, shopping malls, restaurants and even some personal homes. While obtaining video surveillance data is readily supported by existing imaging facilities, manually annotating these data is a time consuming process and it is prone to errors. The noise of data will undoubtedly influence the accuracy of models eventually.
\end{itemize}
\subsection{Relevant Surveys}
To tackle the challenges mentioned above, various algorithms have been devised and gained remarkable experimental results. There have been relevant surveys introducing video anomaly detection models. Kiran et al.~\cite{kiran2018overview} reviewed unsupervised and semi-supervised video anomaly detection models. Mabrouk and Zagrouba~\cite{mabrouk2018abnormal} detailed the procedure inside an intelligent video anomaly detection system, including feature extraction and description. Pawar and Attar~\cite{pawar2019deep} analysed deep learning techniques for video-based anomalous activity detection. Yao and Hu \cite{yao2021survey} introduced both traditional and deep learning-based approaches for video violence detection. Both \cite{nayak2020comprehensive} and \cite{suarez2020survey} provided a comprehensive survey of deep learning-based models for video anomaly detection with minor differences in classifications, while \cite{nayak2020comprehensive} has an additional part evaluating performance among the models. Su et al. \cite{su2021deep} summarised the latest methods of violence detection in existing video sequences. Roshan et al. \cite{roshan2020violence} reviewed the recent trends in violence detection, and performed a comparative study of different state-of-the-art shallow and deep models. Ramzan et al. \cite{ramzan2019review} reviewed various state-of-the-art techniques of violence detection, which is not limited to deep learning models. In \cite{mohammadi2021image}, the authors conducted an in-depth analysis of the deep learning based anomaly detection methods for image and video data. Moreover, current challenges and future research directions are also discussed.

Our work is different from these previous studies in two ways. On the one hand, this survey investigates various applications that video anomaly detection systems could be applied, which is not limited to a fixed domain. On the other hand, we systematically summarise potential opportunities in different applications, and challenges that are still existed in present algorithms, rather than compare the mechanisms behind algorithms as other surveys do.

\subsection{Contributions}
Our contributions are outlined as follows.
\begin{itemize}
    \item A prospective summarization of the opportunities and challenges on video anomaly detection with deep learning methods is presented.
    \item A series of potential research and development directions of intelligent video anomaly detection systems in various application domains have been put forth.
    \item A thorough analysis of major technical challenges in deep learning methods for video anomaly detection has been summarised, thus providing insights into further improvements of models.
\end{itemize}

\subsection{Organization}
The rest of this survey paper is organised as follows. In section~\ref{sec2}, we point out the potential opportunities for deep video anomaly detection models in different real application scenarios. Then, we review existing technical challenges, as shown in section~\ref{sec3}. Finally, the concluding remark of this survey is presented in section~\ref{sec4}.

\section{Opportunities}~\label{sec2}
Most existing studies are devoted to detecting anomalies in traffic video surveillance, while video anomaly detection tasks broadly exist in various real-world scenarios. In this section, we not only introduce the deep video anomaly detection in intelligent transportation, but also outlined several potential opportunities in other fields, i.e., digital education, smart home, public heath, and digital twins.

\subsection{Intelligent Transportation}
Transportation is an essential part of the production, life, and economic development of human society. The current transportation system provides people with fast, comfortable, and safe transportation services~\cite{Ning2019TIIjoint}. However, the increasing demand for transportation by the rapidly growing population have directly led to an explosive increase in the number of motor vehicles. Therefore, problems such as traffic congestion and frequent traffic accidents will follow. In response to this, intelligent transportation system (ITS) consequently came into being, and practice has proved that ITS is an ideal solution to the traffic problems caused by the current economic development~\cite{zhu2018big,Xia2018ICM}.

It is acknowledged that ITS is the most popular research direction among other video anomaly detection applications, which has gained remarkable improvement on detection results as well. Anomaly detection tasks in road traffic scenarios are usually broad, focusing on entities such as vehicles, pedestrian, environment and their interactions~\cite{chen2019survey,kong2018lotad}. Considering that the detection accuracy of traffic monitoring system is influenced by several factors, such as weather and traffic condition, much efforts have been devoted to studying the robustness of detection results in the intelligent traffic system~\cite{el2019robust,Wang2020TITS}.

With recent development of deep learning and wireless communication technologies, numerous innovative traffic monitoring systems have been developed~\cite{chen2021edge,ahmad2021traffic,Xia2019TITSranking}. Li et al.~\cite{li2020multi} aim to detect vehicle anomalies (e.g. traffic accidents) in an unsupervised way. The detection framework is built using a Faster R-CNN \cite{ren2015faster}, which adopts SENet \cite{hu2018squeeze} as the backbone feature extractor. Aboah~\cite{aboah2021vision} proposed a vision-based system for traffic anomaly detection. The anomaly detection process is composed of three main parts: a background estimator to extract background features, a road mask extractor to filter out false anomaly candidates, and a decision tree to confirm and finalise detection results. Despite the consistent development of new deep learning based model for improving the video anomaly detection accuracy in different kinds of environments, there are many open opportunities to be studied in the future work. For example, there is still a huge gap between learning algorithms and real-world deployment of systems. Moreover, the realism of simulation environment in autonomous driving should be improved to ensure the robustness of the model in unstable traffic situations.

\subsection{Online Education}
The traditional offline teaching and learning process is gradually shifting to online platforms due to the development of ICT technologies during the last decade. The outbreak of COVID-19 accelerated this process. Due to this epidemic, online education will be the main way of knowledge delivery within the next period of time. Meanwhile, online examinations popularize as the time requires. Detecting cheating behaviours and proctoring remote online examinations efficiently and effectively is an essential precondition to ensure fairness among examinees ~\cite{stapleton2021remote}.
However, traditional cheating detection methods may no longer be wholly successful in fully preventing cheating during examinations.
It is imperative to devise an artificial intelligence enabled system to automatically detect cheating behaviours during an examination.

\begin{figure*}[htb]
    \centering
    \includegraphics[width=0.99\textwidth]{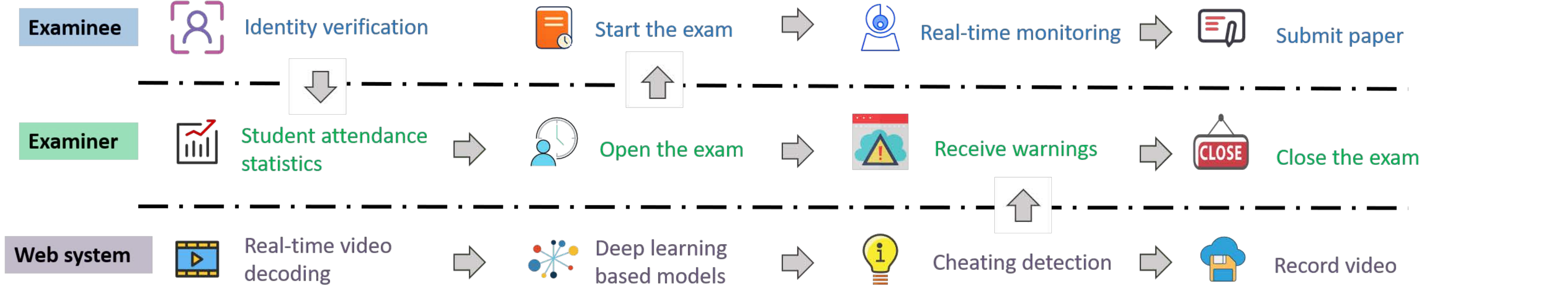}\\
    \caption{The flow diagram of an intelligent invigilation system.}\label{education}
\end{figure*}

Actually, a series of techniques have been developed and applied to intelligent proctoring systems, such as gaze tracking, voice detection, and identification of any entities that are not allowed to exist during an examination. These technologies bring fair and objective examination supervision while saving manpower.
Atoum et al.~\cite{atoum2017automated} proposed an OEP system to automatically and continuously detect cheating behaviours during online exams by using a wearcam and a webcam. Despite that a wearcam could provide a more broad view, equipping every student with a wearcam at home is still not realistic. Bawarith et al. \cite{bawarith2017exam} proposed an online protector in the e-exam management system that realised fingerprint authentication, and eye movement tracking. Moreover, students who move away from the screen can also be detected. Tiong and Lee\cite{tiong2021cheating} presented a deep learning system, DenseLSTM, as the behaviour detection agent. This method can extract better feature representation and strengthen the feature activation of the network, which is effective for predicting potential e-cheating behaviours. A flow diagram of an intelligent invigilation system is shown in Figure~\ref{education}.

In essence, an educational video surveillance system is a complete record of student learning behaviour.
This video data retains more detail than traditional forms of educational data storage. 
For example, for most education stakeholders, including researchers, the score of a course or a student's Grade Point Average (GPA) is often used to evaluate that student's knowledge mastery. This approach brings convenience while losing too much information. With the increase in computing power, we have the ability to process large amounts of data quickly. Recording the learning process through video certainly provides a great help in the analysis of teaching and learning. The video recording of the learning process undoubtedly preserves the entire learning process of the students as well as the examination process. In addition to cheat detection, this provides data security for all education-related anomaly analysis, including course failure analysis, psychological issues, etc.

\subsection{Smart Home}
In order to ensure the safety of the home, many people install video surveillance system at home. Video monitoring is a small part of home automation systems, and is considered a comprehensive security guarantee~\cite{pandya2018smart}. People can use mobile phone and computer to watch the video and master the real-time home situation anytime and anywhere they want~\cite{yang2019security}. Because it is waste of time and energy staring at the screen all the time, automatically identifying anomalous behaviours and sending the alarm signal immediately is undoubtedly necessary.

Yhaya et al. \cite{YAHAYA2021200} proposed an adaptive system for abnormality detection in human activities. This data-driven system adapts to changes in human behavioural routine and has the capacity of discarding old behavioural patterns through embedding a forgetting mechanism. Withanage et al. \cite{withanage2016fall} investigated applied computer vision to recognise fallen postures with RGB-D imaging, to facilitate robot-based in-situ assistantance of falling accidents in elderly independent living. Markovitz et al. \cite{markovitz2020graph} worked directly on human pose graphs that can be constructed from a video sequence, which will not be influenced by nuisance parameters such as viewpoint or illumination. This unsupervised deep learning model could identify anomalous human behaviours by learning normal behaviours. Similarly, Morais et al. \cite{morais2019learning} also learned regularity in skeleton trajectories by modelling the dynamics and interaction of the coupled features in their model. An advantage of this model is that it can provide explanations of its internal reasoning and the visualization of corresponding factors. This is an important part in deep learning based anomaly detection models.

Existing research mostly focus on video monitoring techniques that can record the video fragment when someone appearing in the webcam, while automated anomaly detection is seldom studied. Increases in the elderly dependency ratio is a common problem faced all over the world, which increased additional burden on governments to fund pensions and healthcare~\cite{lee2019does}. Nevertheless, for people who cannot afford a caregiver or prefer to live alone, if the intelligent video anomaly detection system is installed in their homes, the elderly can live independently and the emergency (e.g. the elderly fell down) could be detected and handled timely~\cite{visutsak2017smart}. Therefore, developing video anomaly detection systems in smart home is significant to enhance the quality and convenience of human life. In fact, this kind of intelligent system can also be installed in the hospital and nursing home to reduce the risk brought by unknown accidents~\cite{amisha2019overview}.

\subsection{Public Health}
Public health is an interdisciplinary field, which is related to a variety of areas including epidemiology, biostatistics, social sciences, etc. Besides, environmental health, community health, behavioural health, mental health, and other important sub-areas are also included in the scope of public health. The main purpose of public health is to improve the quality of human life through the prevention and treatment of diseases. By monitoring cases and health indicators, video anomaly detection can benefit public health from many perspectives. Taking the epidemic entitled coronavirus disease 2019 (COVID-19) as example, in order to avoid the further transmission of infectious disease, intelligent video surveillance system can be applied to detect the anomalous behaviours~\cite{jayashri2021video,hou2020social}. Bhambani et al. \cite{bhambani2020real} proposed a real-time face mask and social distancing violation detection system using YOLO object detection on video footage and images. Zuo et al. \cite{zuo2021reference} developed a deep learning-based pedestrian social distancing detection system, which could be used to analyse the new norm of urban mobility amid the pandemic. Saponara et al. \cite{saponara2021implementing} implemented a real-time, AI-based system for COVID-19, which is composed of a deep learning object detection model in combination with a social distancing calculation algorithm.

Intelligent monitoring systems utilise real-time video information to detect anomalous patterns and perform predictive analytics. The anomaly type is then identified and predefined signals will then be initiated to perform remedial actions. Together with wearable sensors, user-specific behaviour pattern, and indoor environment parameters, residents' health conditions can be monitored and further analysed~\cite{2021smart}. Vision-based ambient assisted living, also denoted as AAL, is developed to improve older and vulnerable people's daily lives. Comparing to environmental or body worn sensors, video anomaly detection technologies are much cheaper, more effective, and easier to be implemented. For example, fall detection methods are developed based on RGB camera, multiple cameras, and depth cameras~\cite{zhang2015survey}. Patient monitoring system is also another important application of video anomaly detection in public health. In hospitals, such system is employed to better observe patients regularly, which can detect abnormal activities in wards including irregular poses, unbalanced walking, bed climbing, etc~\cite{prati2019sensors}. Cattani et al.~\cite{2017Monitoring} presented a method to evaluate the periodicity possibility of pathological movements by extracting and processing motion signals from videos. Giving the credit to low cost of camera and the maturity of computer vision technologies, anomaly detection in public health will definitely be further developed. Vision-based methods can be combined with other sensor data to improve their robustness and accuracy. 

\begin{figure*}[htb]
    \centering
    \includegraphics[width=0.8\textwidth]{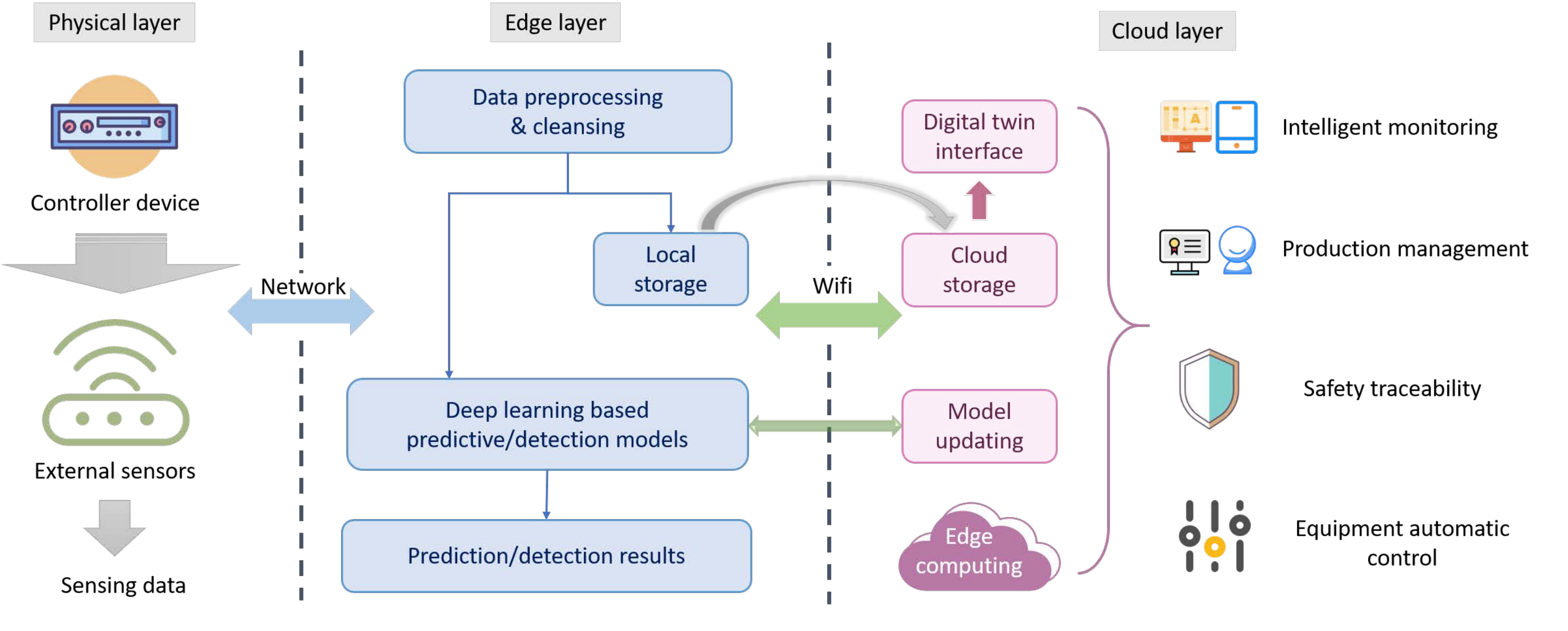}\\
    \caption{An architecture diagram of an anomaly detection/prediction system by combining digital twin technology and deep learning models.}\label{DT}
\end{figure*}

\subsection{Digital Twins}
In an industrial environment, accurate anomaly detection could help the early detection of potential failures and proactive maintenance schedule management~\cite{Yu2020TCSSdetecting}. With the aim of achieving high-performance anomaly detection, recent years have witnessed the growing research interests of implementing Digital Twin technologies in a dynamic industrial edge/cloud network~\cite{huang2021digital}. In general, Digital Twin (DT) technology is used to build a virtual environment that serves as the real-time digital counterpart of a physical object or process. Moreover, advances in Digital Twin technology can help realistic simulations of complex machinery, and thereby speeding up the process of realizing smart manufacturing and Industry 4.0.

Nowadays, the importance of combining digital twin with deep learning is increasingly recognised by both academia and industry in anomaly detection tasks~\cite{he2018surveillance}. In \cite{castellani2020real}, the authors used DT to generate a large dataset of normal operation data covering a complete year of operation. Then, a Siamese Autoencoder (SAE) architecture is applied into anomaly detection in a weakly-supervised way. Due to the critical nature of the power grid, the ability to detect anomalies in the power grid is of critical importance \cite{danilczyk2021smart}. In this paper, the authors used convolutional neural network (CNN) within the Automatic Network Guardian for ELectrical systems (ANGEL) Digital Twin environment to detect physical faults in a power system. The proposed method could not only detect the fault in the power system, but also have the capacity of identifying which bus contains the anomaly. Gao et al. \cite{yiping2021deep} used DT to collect real-time data and realised real-time defect recognition. With the emergence of the new types of anomalies, the traditional models are time-consuming and costly, which have to be rebuilt. To solve this problem, they proposed a deep lifelong learning method for novel class recognition.

It should be noted that all DT-driven anomaly detection systems mentioned above cannot be directly applied to video surveillance data. In modern industries, cameras are deployed with a high density to seamlessly monitor the status of machines and the activities of workers~\cite{he2018surveillance}. DT technology can adopt modern data visualization schemes such as virtual reality (VR) and augmented reality to provide more illustrative and user-friendly views. Therefore, integration of deep learning models and Digital Twin techniques can be further exploited to solve video anomaly detection tasks. Moreover, DT technology has the capacity to generate synthetic data sets including anomalies in different contexts, which solves the problem of lacking dataset with enough positive samples and without noise. An architecture diagram of an anomaly detection/prediction system by combining digital twin technology and deep learning models is shown in Figure~\ref{DT}.

\section{Challenges}~\label{sec3}
Numerous deep learning-based models and intelligent systems have been proposed for the different types of anomalies and technical difficulties encountered in various applications. Obviously, these models and systems can help to reduce human resources consumption to a large extent, and make people's lives more convenient. However, there are still many problems and challenges in video anomaly detection.

In this section, we discuss the technical problems and challenges existed in models according to the model structure (i.e., Reconstruction-based models, Predictive models, Generative models, One-Class classification models, and Hybrid models). There are some connections between the models of different categories. For example, a predictive model can use a generator to predict the next frame of video, and a discriminator to discriminate whether the prediction is real or fake~\cite{nguyen2020anomaly,dong2020dual}. A comparison among these models is summarised in Table~\ref{challenge}.

\begin{table*}
    \label{challenge}
    \caption{The introduction and comparison among different deep video anomaly detection models.}
    \begin{tabular}{|p{1.7cm}<{\centering}|p{8.6cm}<{\centering}|p{4.9cm}<{\centering}|p{1.1cm}<{\centering}|}
        \hline
        \textbf{Type}&\textbf{Assumption}&\textbf{Drawback}&\textbf{Ref.}\\
        \hline
        Reconstruction-based models& Normal data obtains lower value of reconstruction error. Conversely, abnormal data obtains higher value of that & Being invalid when model generalise well; Inexplicability &\multirow{2}{*}{\cite{zhou2017anomaly,ribeiro2018study}}\\ \hline 
        Predictive models   & Normal data can be well predicted, i.e., having closer difference between predicted frame and real frame than abnormal data &\multirow{2}{*} {Higher computational complexity} &\multirow{2}{*}{\cite{chen2020anomaly,deepak2021residual}}\\ \hline
        Generative models&Generator generates irregularities for Discriminator network, and Discriminator is trained as a binary classifier & Expensive training; Instability; Difficulties in reproduction; Mode collapse &\multirow{2}{*}{\cite{an2015variational,wang2018generative}}\\  \hline  
        One-Class classification models& Normal data are compacted into a hyperplane or hypersphere, and anything deviating significantly from the normal behaviour is termed as anomalous   &\multirow{3}{*}{Training with more hours} &\multirow{3}{*}{\cite{sabokrou2018adversarially,wu2019deep}}\\ \hline
        \multirow{3}{*}{    Hybrid models}& Deep learning models are used as feature extractors to generate representations, and representations are input to an classification algorithm & Suboptimal detection performance due to the separation between representation learning and classification model &\multirow{3}{*}{\cite{erfani2016high,zhou2019anomalynet}}\\        \hline
    \end{tabular}
\end{table*}

\subsection{Reconstruction-based Models}
The anomalous instances are often scarce compared with normal instances. To address this problem, reconstruction-based anomaly detection methods usually learn the features of normal behaviours in an unsupervised way. The basic idea of reconstruction model is to reconstruct normal data with low value of reconstruction error in testing phase, and make their distribution closer to training data. Correspondingly, the reconstruction error of anomalous data is expected to be higher. Deep AutoEncoder~\cite{su2019daen} is the most common used model in reconstruction models, which is composed of an \textbf{Encoder} to compress the input vector into a low-dimension embedding, and a \textbf{Decoder} to reconstruct this dense vector back to the input vector. The objective of a DeepAD~\cite{buda2018deepad} is to minimise the reconstruction error $\mathcal{L}$ between the input vector $x_i$ and the reconstructed vector, which could be expressed as:
\begin{equation}
\mathcal{L}=\sum_{i\in \mathrm{N}} \parallel x_i-\mathrm{D}(\mathrm{E}(x_i))\parallel_2,
\end{equation}
where $\mathrm{N}$ is the normal training data, and $\mathrm{D}(\mathrm{E}(\cdot))$ is the DeepAD framework. Here, the Encoder could be any kind of neural networks, such as Convolution Neural Network (CNN), and Long Short-Term Memory (LSTM).

Despite the popularity of DeepAD and its variants, Gong et al.~\cite{gong2019memorizing} pointed out that the assumption of anomaly with higher value of reconstruction error will not be satisfied if an autoencoder is unable to generalise abnormal data. In other words, the anomaly is reconstructed using a generalised model, and the generated representation by encoder cannot guarantee its validity. Thus, the model cannot explain why the detected anomaly frame is anomalous.

\subsection{Predictive Models}
Video is composed of a series of frames, which could be viewed as an order of spatial and temporal signals. The task of a predictive model is to predict the $t$ frame by giving the past $p$ frames, which could be expressed as:
\begin{equation}
x^{'}_t=\mathrm{h}(x_{t-1},x_{t-2},...,x_{t-p}).
\end{equation}
The loss function of a predictive model is constructed based on the real target frame and its prediction frame:
\begin{equation}
\mathcal{L}=\sum_{t=1}^{m}\parallel x_t-x_t^{'} \parallel_2^2,
\end{equation}
where $x_t$ is the real target frame in timestamp $t$, and $x_t^{'}$ is the predicted frame. The predictive model assumes that normal events can be well predicted. Therefore, the difference between predicted frame $x_t'$ and its ground truth $x_t$ can be used to detect abnormal events.
Although the predictive models perform well in video anomaly detection tasks, it has higher computational complexity. Therefore, predictive model is more suitable for offline applications.

\subsection{Generative Models}
Generative models usually contain an architecture to generate frames based on Gaussian distribution, such as Generative adversarial networks (GAN)~\cite{goodfellow2020generative}. GAN is composed of a generator and a discriminator. The role of the generator is to fit a new data distribution according to the actual distribution of the real data, and the discriminator is to discriminate whether the vector is extracting from real data or generated data. The loss function of GAN is expressed as follows:
\begin{equation}
\mathcal{L}=\frac{1}{m}\sum_{i=1}^{m}[{\mathrm {log}} \mathrm{D}(x_i)+{\mathrm {log}}(1-\mathrm{D}(\mathrm{G}(z_i)))].
\end{equation}
The front part of this function aims to maximise the probability of identifying the real data, and the latter is to discriminate the generated data. Here, generator and discriminator could be any kind of neural network architectures, like CNN. Different from other models, GAN could serve as an end-to-end model by simultaneously training the generator and the discriminator. Moreover, the generator could generate the abnormal samples at the same time. Therefore, GAN is one of the most widely used models in video anomaly detection. Despite its advantages, GAN suffers from some inevitable defects, including expensive training, instability, difficulties in reproduction and mode collapse.

\subsection{One-Class Classification Models}
Considering the ambiguity and diversity of anomaly, the development of multi-class classification for the detection of video anomaly is urgently needed. In detection of video anomaly, researchers often treat anything deviating significantly from the normal behaviour to be termed as anomalous. Thus, the anomaly detection task with no anomalous labels could be viewed as a one-class classification (OCC) problem. The core idea of this kind of model in video anomaly detection is to find a hypersphere that encloses the network representations of the normal data~\cite{ruff2018deep}. Any data points that are not included in this hypersphere will be considered anomalous. The combination of deep learning and OCC models could be trained to learn the dense feature representation with the one-class classification objective jointly. However, this kind of model requires extended training time~\cite{pang2021deep}.

\subsection{Hybrid Models}
Every kind of model has its own objective function and specified advantage in solving anomaly detection tasks. Therefore, researchers can consider making multiple models serving different blocks in one model, which could take advantage of different models and improve the detection accuracy. In the hybrid models, the learned representative features from deep learning methods can be transferred to traditional algorithms like Support Vector Machine (SVM) classifiers~\cite{chalapathy2019deep}. The low dimensional feature vectors make hybrid models more scalable and computationally efficient, which is suitable for solving video anomaly detection tasks. Different from other models that have customised loss function, the loss function of a hybrid model is generic, which means the feature extractor has no influence on the feature representation. As a result, the performance of the hybrid model is suboptimal. Even though the hybrid Models have excellent performance in tasks, they are mostly task-dependent and not able to switch between different tasks. 

\section{Conclusion}~\label{sec4}
In this paper, we presented the potential opportunities for deep video anomaly detection models in several emerging real-world application scenarios, and discussed the technical problems in the literatures. The novel perspective of this survey in deep video anomaly detection offers a clear guidance to researchers who are interested in this field.

\section*{Acknowledgment}
The authors would like to thank Teng Guo, Shuo Yu, and Ke Sun for their help with the first draft of this paper.

\bibliographystyle{IEEEtran}

\bibliography{BIB}
\end{document}